%% file: main.tex
\documentclass{article}
\usepackage{preprint,times}
\usepackage{amsmath,amsfonts}
\usepackage{hyperref}
\usepackage{url}

\usepackage{graphicx}
\usepackage{booktabs}
\usepackage{multirow}
\usepackage{marvosym}
\pdfglyphtounicode{Letter}{2709}

\title{Constructing Pareto Frontiers of Optimal Reasoning Termination Methods for Overthinking in Reasoning Models}

\author{\begin{minipage}[t]{\textwidth}
\normalfont\centering
{\bfseries
Weixin Guan\textsuperscript{1,2}\quad
Liang Li\textsuperscript{1,}\thanks{Corresponding author: Liang Li.}\quad
Shiyu Xie\textsuperscript{3}\quad
Jiapeng Liu\textsuperscript{1,2}\quad
Bing Li\textsuperscript{1}\\[2pt]
Peng Fu\textsuperscript{1}\quad
Chengyang Fang\textsuperscript{4}\quad
Xiaoshuai Hao\textsuperscript{5}\quad
Can Ma\textsuperscript{1}\quad
Weiping Wang\textsuperscript{1}\endgraf}
\vspace{0.5em}
{\small
\textsuperscript{1}\,Institute of Information Engineering, Chinese Academy of Sciences, Beijing, China\\[1pt]
\textsuperscript{2}\,School of Cyberspace Security, University of Chinese Academy of Sciences, Beijing, China\\[1pt]
\textsuperscript{3}\,University of Florida, Gainesville, FL, USA\\[1pt]
\textsuperscript{4}\,School of Computer and Artificial Intelligence,
Jiangxi University of Finance and Economics\\[1pt]
\textsuperscript{5}\,Li Auto Inc.\endgraf}
\end{minipage}}

\hypersetup{
  pdftitle={Constructing Pareto Frontiers of Optimal Reasoning Termination Methods for Overthinking in Reasoning Models},
  pdfauthor={Weixin Guan, Liang Li, Shiyu Xie, Jiapeng Liu, Bing Li, Peng Fu, Chengyang Fang, Xiaoshuai Hao, Can Ma, Weiping Wang}
}

\begin{document}

\maketitle

{\let\thefootnote\relax\footnotetext[0]{Email: Weixin Guan\textless\href{mailto:guanweixin@iie.ac.cn}{guanweixin@iie.ac.cn}\textgreater,
Liang Li\textless\href{mailto:liliang@iie.ac.cn}{liliang@iie.ac.cn}\textgreater.}}

\begin{abstract}
Reasoning models can benefit from test-time scaling, but continued reasoning may become ineffective or harmful, a phenomenon termed overthinking.
Evaluations against full thinking compare accuracy and token usage but lack references for overthinking severity and strategies’ theoretical mitigation potential.
To characterize overthinking at the model--task level, we first introduce Step-wise Truncation Probing (STP), repeatedly generating answers from multiple retained prefixes along each fixed sampled reasoning trajectory.
Using each trajectory’s earliest measured position attaining maximum empirical accuracy as a reference, we define Overthinking Ratio and Accuracy Degradation as the fraction of thinking tokens generated afterward and the accuracy decrease from this reference to full thinking, respectively.
Secondly, we then formulate dynamic routing, fixed budgets, and dynamic early exit within a unified framework to quantify the theoretical mitigation potential of different termination strategies.
Based on this unified framework, we use the complete STP measurements to construct a token–accuracy Pareto frontier for each strategy under a common normalized thinking-token budget.
We further develop OTBench, a reusable benchmark built from the STP process, which produced 70.5B reasoning tokens and the corresponding probing measurements across six models and four benchmark groups.
\end{abstract}

\section{Introduction}

Reasoning models can improve their performance on complex tasks through test-time scaling~\citep{muennighoff2025s1,xu2025scalable}.
However, continued reasoning does not always improve answer quality and can even degrade it~\citep{chen2025overthinking,su2025between,wu2025more,gema2025inverse}.
Such ineffective or harmful continued reasoning is referred to as \emph{overthinking} ~\citep{zhou2026thinkinghurtsoverthinkingllm,duan2026circularreasoningunderstandingselfreinforcing}.
To achieve a trade-off between performance and efficiency, recent methods mitigate overthinking by controlling when reasoning ends.
Based on when this termination decision is made, these methods can be organized into three strategies.
Dynamic routing~\citep{ma2026odarprincipledadaptiverouting,liu2026diffadaptdifficultyadaptivereasoningtokenefficient} decides, before generation, whether a question should trigger reasoning.
Fixed budgets~\citep{brown2026predictive,lin2026planbudgeteffectiveefficient} specify a thinking-token cap before generation.
Dynamic early exit~\citep{guan2026rpdi,yang2025dynamic} decides, during generation, whether to terminate the current reasoning prefix.

Existing termination methods are typically evaluated against full thinking by comparing accuracy and token usage~\citep{li2025thinkbench,aggarwal2025optimalthinkingbench,tan2025zero}. 
Such comparisons only show the relative gain resulting from the termination points selected by a given method, i.e., whether it outperforms full thinking and other comparable methods. 
However, they lack reference points for both the degree of overthinking and the best token--accuracy trade-offs theoretically available to a given termination strategy under idealized choices.
Consequently, they cannot answer two key questions for a model on a given task type: 
(1) how severe is its overthinking, and 
(2) how much can different termination strategies theoretically mitigate overthinking.
This lack of reference points hinders the field's ability to assess the potential of termination strategies and to determine whether reported gains reflect substantial mitigation of overthinking.

\begin{figure}[t]
\centering
\includegraphics[width=\linewidth]{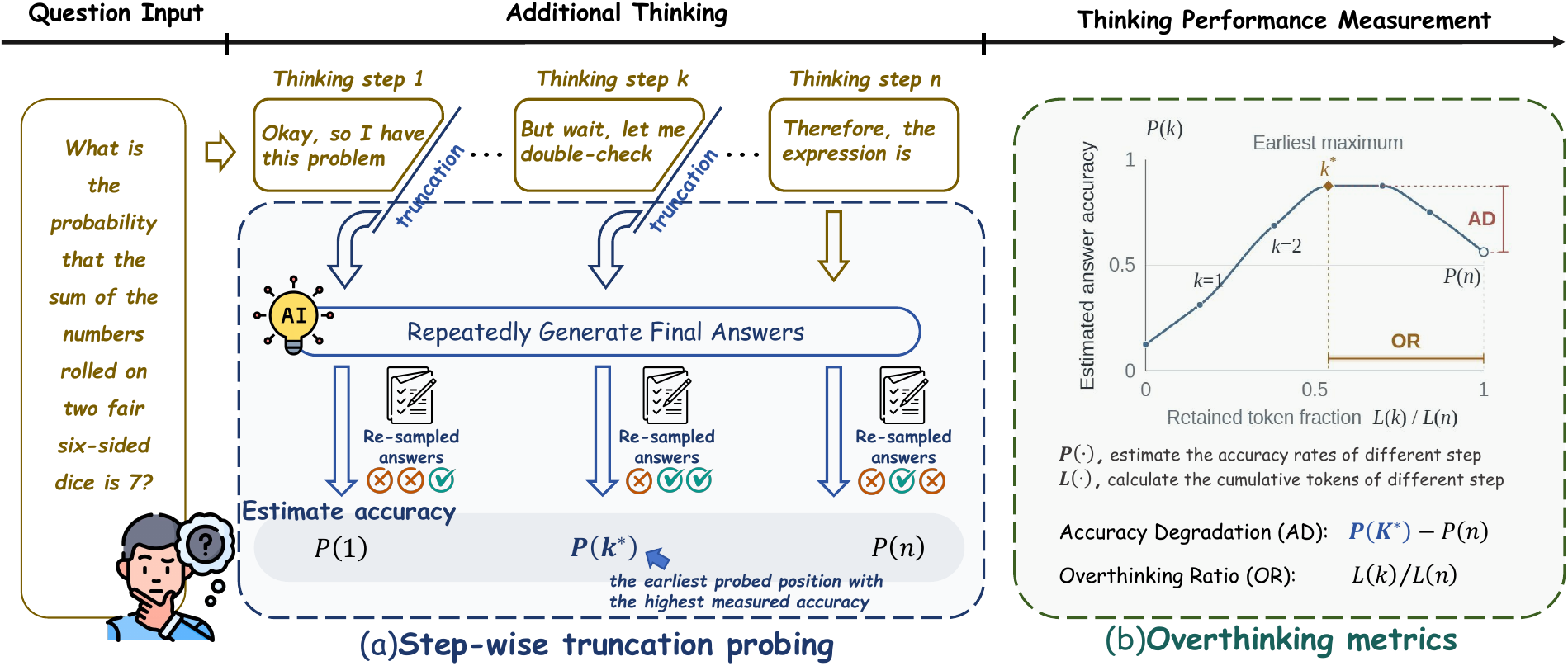}
\caption{
\textbf{Step-wise truncation probing and overthinking metrics.}
(a) STP measures the corresponding accuracy by terminating reasoning at different points along the same reasoning trajectory and repeatedly generating answers from each retained prefix.
(b) OR denotes the proportion of reasoning length after accuracy first reaches its maximum. AD denotes the decrease in accuracy under full reasoning relative to the maximum accuracy.
}
\label{fig:stp-overview}
\end{figure}

In this paper, to answer the two questions above, we proceed in two steps.
\textbf{We first introduce \emph{Step-wise Truncation Probing} (STP) to measure the impact of continued reasoning and characterize overthinking at the model--task level}, as illustrated in Figure~\ref{fig:stp-overview}.
Specifically, STP probes multiple truncation positions along the same fixed sampled reasoning trajectory and repeatedly generates answers conditioned on each retained prefix.
This produces a token--accuracy response curve for each trajectory while holding the generated reasoning path fixed, providing trajectory-level measurements that can be aggregated across sampled trajectories.
For each trajectory, the earliest measured position attaining the highest empirical answer accuracy is used only as a trajectory-level reference point for constructing OR and AD, not as a common stopping point across trajectories.
From this reference point, we define the \emph{Overthinking Ratio} (OR) as the fraction of thinking tokens generated after the reference point, and the \emph{Accuracy Degradation} (AD) as the decrease in answer accuracy from the reference point to full thinking.
We aggregate these trajectory-level measurements across sampled trajectories to characterize overthinking at the model--task level.

\textbf{We then use the complete STP measurements to quantify the theoretical potential of the three termination strategies.} 
Specifically, under a common normalized thinking-token budget, we formulate dynamic routing, fixed budgets, and dynamic early exit in a unified framework and construct a token--accuracy Pareto frontier for each termination strategy.
These frontiers represent the best admissible token--accuracy trade-offs under the measured action space and each strategy's decision process, and serve as idealized theoretical references rather than deployable stopping algorithms. 
To connect these references with practical methods, we further construct \emph{OTBench}, a benchmark for comparing termination methods with the corresponding strategy-specific token--accuracy Pareto frontier.
\emph{OTBench} turns the STP measurements and strategy-specific frontiers into a common evaluation protocol with public data splits, a unified method interface, and a dedicated evaluator.
Using OTBench, we evaluate one publicly available method from each strategy against its corresponding token--accuracy Pareto frontier, thereby assessing how much of each strategy's theoretical potential the selected methods realize.

Experimental results on six models spanning different sizes and families across four benchmarks reveal three findings:
(1) \textbf{overthinking is substantial and varies across trajectories}, with 64.3\% of thinking tokens generated after the highest measured accuracy is first attained and full thinking incurring a 9.03-percentage-point accuracy drop relative to the reference point.
(2) \textbf{under a common normalized thinking-token budget, the three frontiers exhibit a clear hierarchy}: the dynamic-routing frontier lies no higher than the fixed-budget frontier, which lies no higher than the dynamic-early-exit frontier. This ordering shows that the theoretical mitigation potential increases with the flexibility of the strategy's decision process.
(3) the selected public baselines lie below their corresponding frontiers, indicating substantial unrealized space relative to the theoretical potential of their respective strategies.

The main contributions of this paper are as follows:
\begin{itemize}
    \item We propose \emph{Step-wise Truncation Probing} (STP), which provides a quantitative characterization of overthinking at the model--task level, revealing it to be substantial and trajectory-dependent.
    \item We construct strategy-specific token--accuracy Pareto frontiers for the three termination strategies under a common normalized thinking-token budget, providing idealized theoretical references for their termination potential.
    \item We propose \emph{OTBench}, a public, reusable benchmark for evaluating termination methods against their corresponding strategy-specific token--accuracy Pareto frontiers.
\end{itemize}

\section{Related Works}

Research on overthinking has documented redundant computation, diminishing benefits, and performance degradation under particular reasoning conditions.
\citet{chen2025overthinking} analyze complete responses and identify redundant reasoning after a solution or answer has appeared, particularly on simpler problems.
Comparing independently sampled trajectories ordered by length, \citet{su2025between} observe a nonmonotonic relationship between reasoning length and correctness, while \citet{wu2025more} study how accuracy varies with chain-of-thought length across task difficulties and model capabilities.
\citet{gema2025inverse} and \citet{ghosal2025mirage} likewise report performance degradation under increased reasoning budgets on selected tasks and settings.
Other studies examine internal bias and missing premises as sources of reasoning failure~\citep{dang2026first,fan2025missing}, as well as underthinking and the possibility of answering without explicit thinking~\citep{wang2025underthinking,ma2025nothinking}.

Overthinking mitigation methods include approaches that modify reasoning generation and approaches that select when reasoning ends.
ThinkPrune, for example, uses reinforcement learning to encourage shorter reasoning chains~\citep{hou2025thinkprune}.
Related training-time approaches explicitly optimize reasoning efficiency or control the length of generated reasoning, thereby changing the trajectories a model produces~\citep{arora2025efficient,aggarwal2025lengthcontrol}.
In contrast, termination approaches~\citep{liang2025thinkswitcher,fang2025thinkless} control how much of a generated trajectory is retained before answering.

Within termination methods, we distinguish dynamic routing, fixed budgets, and dynamic early exit according to when the termination decision is made.
Dynamic routing uses question-level information before generation to decide whether to invoke reasoning or which reasoning mode to use; ThinkSwitcher predicts the relative benefit of short and long chain-of-thought reasoning, while other routing methods learn when to think or select a hybrid reasoning mode~\citep{liang2025thinkswitcher,fang2025thinkless,jiang2025thinkonly,lu2025prolonged}.
Fixed budgets specify a thinking-token cap before generation. Predictive Scheduling predicts success under candidate reasoning budgets to allocate computation across questions, while budget-aware reasoning and elastic chain-of-thought methods make thinking computation an explicit control variable~\citep{brown2026predictive,han2025token,muennighoff2025s1,xu2025scalable}.
Dynamic early exit decides during generation whether to terminate the current reasoning prefix. ThinkBrake uses token log-probabilities at permitted stopping positions~\citep{song2025thinkbrake}, while other methods use answer probes, confidence or hidden-state signals, entropy, learned stopping rules, or reasoning-path deviation~\citep{fu2025certaindex,yang2025dynamic,wang2025eat,sun2025stop,akgul2025lynx,guan2026rpdi}.

\section{Characterizing Overthinking at the Model--Task Level}

We introduce Step-wise Truncation Probing (STP) as a measurement procedure for characterizing overthinking at the model--task level.
STP measures answer accuracy at different retained thinking-token counts along the same reasoning trajectory and identifies the shortest measured prefix attaining the highest measured answer accuracy as a trajectory-level empirical reference.
We quantify overthinking using Overthinking Ratio (OR) and Accuracy Degradation (AD), and evaluate six reasoning models across four benchmark groups.
Beyond its overall magnitude, we examine whether length redundancy tracks accuracy loss, how local accuracy changes evolve, and how overthinking varies across trajectories for the same model and question.

\subsection{Measuring Overthinking with Step-wise Truncation Probing}

\textbf{Probing a fixed trajectory at multiple stopping positions.}
STP partitions a generated reasoning trajectory into steps and repeatedly generates answers after truncating the trajectory at each step boundary (Figure~\ref{fig:stp-overview}).
We form initial text blocks at blank-line boundaries and repeatedly merge the adjacent pair with the smallest combined token count until reaching a target step count determined by the trajectory length and target average step length $b$.
For a trajectory with $n$ steps, probing position $k\in\{0,\ldots,n\}$ retains the first $k$ steps; only the answer is resampled, while the reasoning prefix remains fixed.
Let $L(k)$ be the retained thinking-token count and $P(k)$ the fraction of correct answers at that position.
The pairs $(L(k),P(k))$ form the trajectory's measured token--accuracy curve, including no thinking at $k=0$ and the complete trajectory at $k=n$.

\textbf{Quantifying length redundancy and accuracy degradation.}
We select the earliest measured position attaining the highest measured answer accuracy:
\begin{equation}
k^\star = \min\left\{k : P(k)=\max_{0\leq j\leq n} P(j)\right\}.
\end{equation}
Relative to this reference, we measure the additional thinking-token count and the accuracy difference associated with continuing to the STP full-thinking endpoint:
\begin{equation}
\mathrm{OR}=1-\frac{L(k^\star)}{L(n)},
\label{eq:overthinking-ratio}
\end{equation}
\begin{equation}
\mathrm{AD}=P(k^\star)-P(n).
\label{eq:accuracy-degradation}
\end{equation}
OR measures the fraction of thinking tokens generated beyond $k^\star$, while AD measures the accuracy difference between this reference and the STP full-thinking endpoint $P(n)$; we report them in percent and percentage points, respectively.

\subsection{Quantifying overthinking for six models across four benchmark groups}

We use STP to quantify overthinking at the model--task level.
Specifically, we evaluate the six models listed in Table~\ref{tab:main-results} on four reporting benchmark groups: AIME (AIME2024, AIME2025, and AIME2026 combined) ~\citep{aime}, MATH500 (MATH)~\citep{hendrycks2021measuring}, GPQA-Diamond (GPQA)~\citep{rein2024gpqa}, and LiveCodeBench (LCB)~\citep{jain2024livecodebenchholisticcontaminationfree}.
For each model--question pair, we sample 16 reasoning trajectories and set $b=256$ with 16 answer probes per position.
We compute OR and AD per trajectory, average trajectories within each question and questions within each benchmark, then weight the four benchmarks and six models equally.
The original full-thinking baseline accuracy (Table~\ref{tab:appendix-full-thinking} in Appendix~\ref{app:protocol}) uses the original final answers and is distinct from the STP full-thinking endpoint accuracy $P(n)$ estimated from repeated answer probes at $k=n$.
Appendices~\ref{app:protocol} and~\ref{app:uncertainty} provide protocol details and measurement-robustness analyses, while Appendix~\ref{app:curves} provides additional trajectory-level examples.
The resulting model--task-level measurements are summarized in Table~\ref{tab:main-results}.
We summarize these results in four findings:

\textbf{Substantial thinking continues beyond the measured accuracy maximum.}
The original full-thinking baseline has an average answer accuracy of 71.78\% and an average of 8,281 thinking tokens (Table~\ref{tab:appendix-full-thinking}).
The mean OR is 64.3\%, indicating that a substantial fraction of thinking tokens extends beyond the trajectory-level measured reference position, while the STP full-thinking endpoint $P(n)$ is 9.03 percentage points less accurate than that reference on average (Table~\ref{tab:main-results}).
The benchmark-averaged results show substantial overthinking for all six models: OR ranges from 41.1\% to 74.4\%, while AD ranges from 5.64 to 13.40 percentage points.

\begin{table}[t]
\centering
\caption{
\textbf{STP-based overthinking measurements.}
AD is reported in percentage points (pp) and OR in \%.
Avg. uses equal benchmark and model weights.
Models: M1: GPT-OSS-20B; M2: Qwen3-4B-Thinking-2507; M3: Qwen3-30B-A3B-Thinking-2507; M4: DeepSeek-R1-Distill-Qwen-7B; M5: DeepSeek-R1-Distill-Qwen-14B; M6: DeepSeek-R1-Distill-Qwen-32B.
}
\label{tab:main-results}
\begingroup
\fontsize{9}{11}\selectfont
\setlength{\tabcolsep}{1.5pt}
\renewcommand{\arraystretch}{1.12}
\begin{tabular*}{\linewidth}{@{\extracolsep{\fill}}l*{5}{rr}@{}}
\toprule
\multirow{2}{*}{\textbf{Model}} & \multicolumn{2}{c}{\textbf{AIME}} & \multicolumn{2}{c}{\textbf{MATH}} & \multicolumn{2}{c}{\textbf{GPQA}} & \multicolumn{2}{c}{\textbf{LCB}} & \multicolumn{2}{c}{\textbf{Avg.}} \\
\cmidrule(lr){2-3}\cmidrule(lr){4-5}\cmidrule(lr){6-7}\cmidrule(lr){8-9}\cmidrule(lr){10-11}
 & \textbf{AD} & \textbf{OR} & \textbf{AD} & \textbf{OR} & \textbf{AD} & \textbf{OR} & \textbf{AD} & \textbf{OR} & \textbf{AD} & \textbf{OR} \\
\midrule
\textbf{M1} & 3.02 & 34.4 & 1.66 & 45.4 & 16.15 & 35.4 & 9.89 & 49.1 & 7.68 & 41.1 \\
\textbf{M2} & 2.52 & 63.4 & 1.24 & 89.3 & 14.23 & 71.6 & 18.28 & 57.6 & 9.07 & 70.5 \\
\textbf{M3} & 1.53 & 63.5 & 1.73 & 92.0 & 10.69 & 73.9 & 8.63 & 68.2 & 5.64 & 74.4 \\
\textbf{M4} & 9.76 & 62.9 & 2.82 & 82.4 & 25.64 & 58.8 & 15.37 & 62.8 & 13.40 & 66.7 \\
\textbf{M5} & 7.52 & 59.0 & 1.80 & 79.3 & 18.73 & 65.2 & 9.95 & 55.9 & 9.50 & 64.8 \\
\textbf{M6} & 4.96 & 59.9 & 1.75 & 84.6 & 16.42 & 63.6 & 12.40 & 66.2 & 8.88 & 68.6 \\
\midrule
\textbf{Avg.} & \textbf{4.88} & \textbf{57.2} & \textbf{1.83} & \textbf{78.8} & \textbf{16.98} & \textbf{61.4} & \textbf{12.42} & \textbf{59.9} & \textbf{9.03} & \textbf{64.3} \\
\bottomrule
\end{tabular*}
\par
\endgroup
\end{table}

\textbf{Greater length redundancy does not imply greater accuracy degradation.}
Across benchmark groups, after averaging over models, MATH has the highest mean OR (78.8\%) but the lowest AD (1.83 percentage points), whereas GPQA has a lower OR (61.4\%) but the highest AD (16.98 percentage points; Table~\ref{tab:main-results}).
Thus, a larger measured OR need not imply a larger endpoint accuracy loss.
Across the 24 model--benchmark means, the Spearman correlation between OR and AD is $-0.330$, indicating a weak negative association at this aggregate level.

\begin{figure}[t]
\centering
\includegraphics[width=\linewidth]{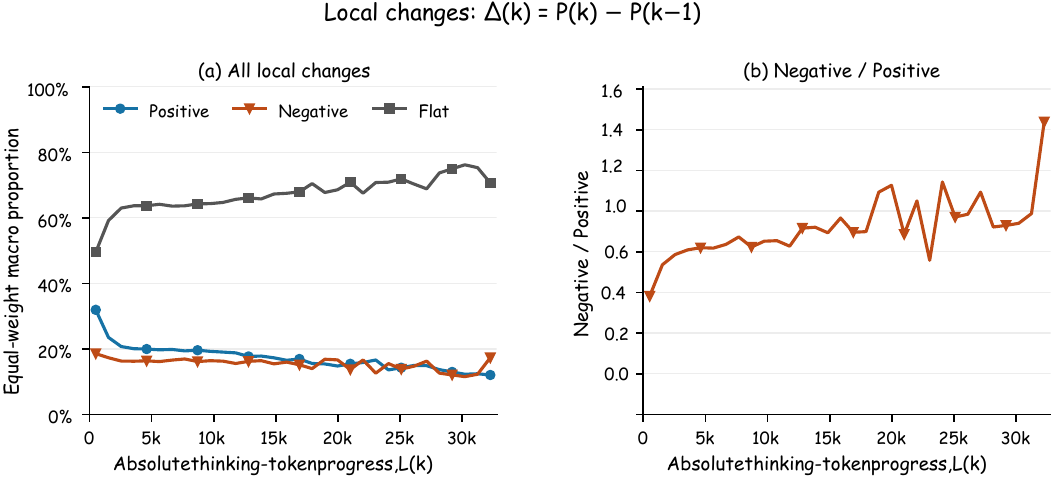}
\caption{
\textbf{Local measured accuracy changes between adjacent STP measurements.}
$\Delta P(k)=P(k)-P(k-1)$.
Left: equal-weight macro-averaged proportions of positive, negative, and flat changes.
Right: the ratio of negative to positive changes.
Both panels use the absolute number of retained thinking tokens $L(k)$ on the horizontal axis.
Later positions include only trajectories long enough to reach them.
The signs reflect finite-probe accuracy differences, not the logical correctness or causal contribution of individual reasoning steps.}
\label{fig:local-accuracy-changes}
\end{figure}

\textbf{Later steps mostly leave accuracy unchanged, with relatively more negative changes.}
Figure~\ref{fig:local-accuracy-changes} plots the proportions of positive, negative, and flat finite-probe accuracy changes, $\Delta P(k)=P(k)-P(k-1)$, together with the negative-to-positive ratio as a function of absolute thinking-token progress $L(k)$.
At early measured positions, positive finite-probe accuracy changes are more common than negative changes, and the flat fraction is relatively low.
At later measured positions, flat changes become more prevalent and the negative-to-positive ratio generally rises, exceeding one in some later measured intervals.
The dominant shift is therefore toward steps with no measured accuracy change, rather than a growing overall fraction of negative changes.
This increasingly flat measured continuation is consistent with the token redundancy captured by OR.
Among the remaining non-flat measured changes, decreases become more common relative to improvements, a pattern consistent with AD.

\textbf{Overthinking varies across trajectories for the same model and question.}
For each model--question pair, we compute the mean absolute difference in OR and AD over the 120 pairs of its 16 trajectories, then aggregate across questions, benchmarks, and models as above.
The average pairwise absolute difference is 20.23 percentage points for OR and 10.06 percentage points for AD.
Even with the model and question fixed, both the measured token redundancy and the endpoint accuracy difference vary substantially across sampled trajectories.

\section{Pareto Frontiers of Oracle Reasoning-Termination Methods}

Building on the STP measurements, we construct offline token--accuracy Pareto frontiers for oracle methods under dynamic routing, fixed budgets, and dynamic early exit.
To do so, we formulate each termination strategy as a budget-constrained action-selection problem over the measured action space.
Because STP associates each measured action with a thinking-token cost and an answer accuracy, the oracle for each strategy can select the feasible actions with the highest measured accuracy at each budget.
The resulting frontier therefore represents the best token--accuracy trade-off available under that strategy's measured action space.
These frontiers provide the theoretical references for comparing termination-strategy potential and budget requirements.

\subsection{A Unified Framework for Oracle Termination Methods}

To construct offline Pareto frontiers for oracle termination methods, we cast dynamic routing, fixed budgets, and dynamic early exit as a unified budget-constrained action-selection problem.
For one model--benchmark pair and a termination strategy $C\in\{\mathrm{DR},\mathrm{FB},\mathrm{DEE}\}$, let $\mathcal{U}_C$ be the set of decision units and $\mathcal A_u^C$ the available actions for unit $u$.
Each action $a$ has a retained thinking-token cost $c_u(a)$ and answer accuracy $r_u(a)$, both obtained from STP.
An oracle method chooses one action per unit to solve
\begin{equation}
\begin{aligned}
F_C(B)
&=\max_{\{a_u\in\mathcal A_u^C\}_{u\in\mathcal{U}_C}}
\frac{1}{|\mathcal{U}_C|}\sum_{u\in\mathcal{U}_C}r_u(a_u),\quad
\text{subject to}
&\frac{1}{|\mathcal{U}_C|}\sum_{u\in\mathcal{U}_C}c_u(a_u)\leq B.
\end{aligned}
\label{eq:oracle-optimization}
\end{equation}To compare oracle termination methods, we cast dynamic routing, fixed budgets, and dynamic early exit as a unified budget-constrained action-selection problem.
The objective averages the accuracy of the selected actions, while the constraint limits their average thinking cost to $B$ tokens.
Changing the decision units and allowed actions instantiates each strategy without changing this objective.

Index questions by $q$ and their $R=16$ sampled trajectories by $s$, with $n_{qs}$ steps, retained length $L_{qs}(k)$, and probe accuracy $P_{qs}(k)$.
Table~\ref{tab:termination-constraints} specifies the decision unit, action, and resulting stopping position for each strategy.
Dynamic routing shares a zero-or-full endpoint choice across a question's trajectories, fixed budgets share a question-level token cap, and dynamic early exit chooses a measured position independently for each trajectory.
For fixed budgets, candidate caps are the union of measured token counts for that question and different questions can receive different caps.

\begin{table}[t]
\centering
\caption{
\textbf{Termination strategies and corresponding actions of oracle methods in the unified framework.}
Here $q$ indexes questions, $s$ indexes sampled trajectories, and $n_{qs}$ is the number of steps in a trajectory.
Question-level actions apply to all sampled trajectories of the same question.}
\label{tab:termination-constraints}
\begingroup
\small
\setlength{\tabcolsep}{4pt}
\renewcommand{\arraystretch}{1.25}
\begin{tabular*}{\linewidth}{@{\extracolsep{\fill}}llp{0.52\linewidth}@{}}
\toprule
\textbf{Strategy} & \textbf{Decision unit} & \textbf{Action and stopping position} \\
\midrule
Dynamic routing (DR) & Question $q$ & Endpoint $a\in\{0,1\}$ (zero/full thinking): $k_{qs}(a)=a n_{qs}$ \\
Fixed budgets (FB) & Question $q$ & Token cap $\tau_q$: $k_{qs}(\tau_q)=\max\{k:L_{qs}(k)\leq\tau_q\}$ \\
Dynamic early exit (DEE) & Trajectory $(q,s)$ & Position $k\in\{0,\ldots,n_{qs}\}$ \\
\bottomrule
\end{tabular*}
\endgroup
\end{table}

For a question-level action $a$, we average its retained lengths and probe accuracies over that question's trajectories:
\begin{equation}
c_q(a)=\frac{1}{R}\sum_{s=1}^{R}L_{qs}\bigl(k_{qs}(a)\bigr),
\label{eq:oracle-question-cost}
\end{equation}
\begin{equation}
r_q(a)=\frac{1}{R}\sum_{s=1}^{R}P_{qs}\bigl(k_{qs}(a)\bigr).
\label{eq:oracle-question-accuracy}
\end{equation}
Here $k_{qs}(a)$ maps the shared action to each trajectory's stopping position, so that a shared cap can retain different token counts on different trajectories.
For a trajectory-level action $k$, the cost and accuracy are simply $c_{qs}(k)=L_{qs}(k)$ and $r_{qs}(k)=P_{qs}(k)$.
Because every question has the same $R$, both decision-unit choices yield the same per-trajectory accuracy and cost scale.

We solve Equation~\ref{eq:oracle-optimization} exactly over deterministic combinations of measured actions, retaining accuracy--cost pairs (Appendix~\ref{app:oracle-computation}).
This gives an offline optimum within each strategy's measured action space and stopping constraints.
To compare the same six models and four benchmark groups, let $g$ index their 24 combinations and $T_g$ denote the mean full-thinking token count in combination $g$.
The macro-average frontier at relative budget $\beta\in[0,1]$ is
\begin{equation}
\overline F_C(\beta)=\frac{1}{24}\sum_{g=1}^{24}F_C^{(g)}(\beta T_g).
\label{eq:oracle-macro-average}
\end{equation}
Each combination thus receives the same fraction of its own full-thinking budget before accuracies are averaged with equal model and benchmark weights.
The horizontal coordinate is an allowed budget cap.
The selected actions may use fewer tokens.

\subsection{Frontier Hierarchy Across Termination Strategies}

\begin{figure}[t]
\centering
\includegraphics[width=\linewidth]{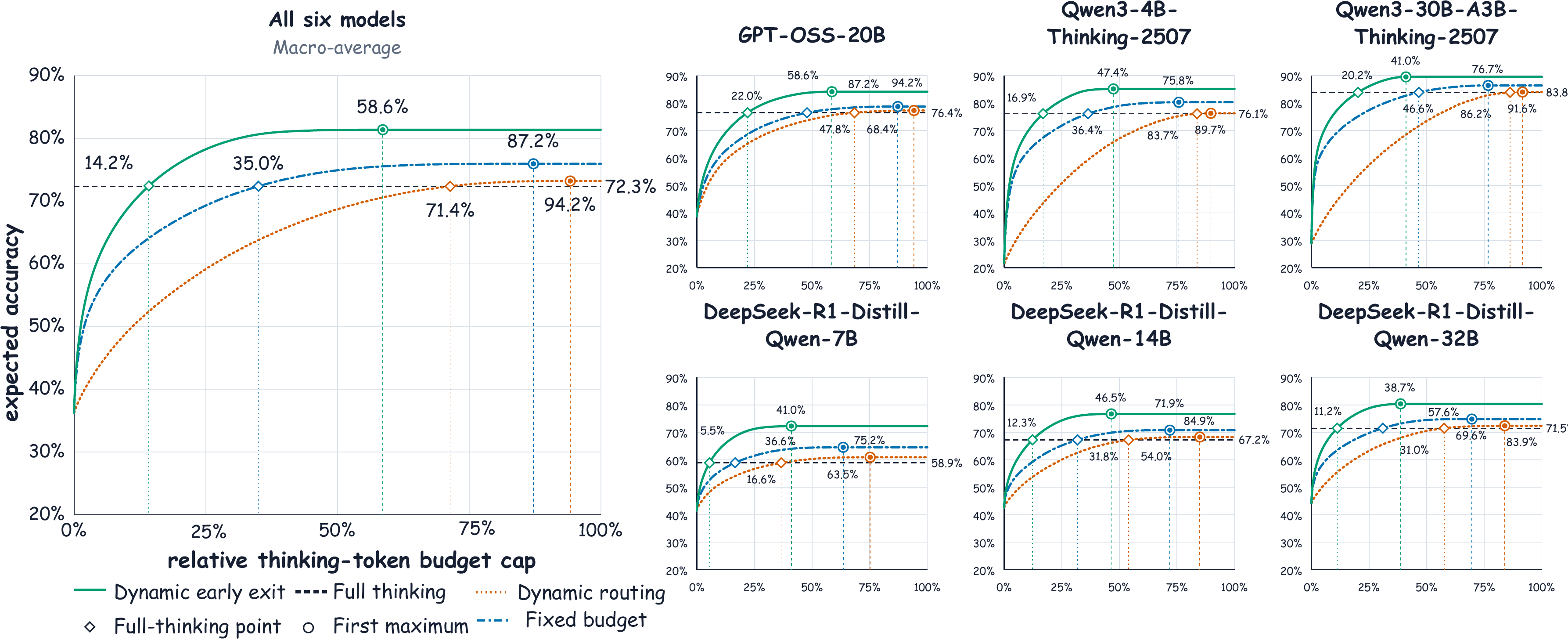}
\caption{
\textbf{Pareto frontiers for oracle reasoning termination.}
Left: averages over six models and four benchmark groups; right: per-model benchmark averages, with equal weights throughout.
The x-axis shows budget caps normalized by full-thinking cost, not realized costs.
Dashed lines mark the macro-average STP full-thinking endpoint accuracy $P(n)$.
Diamonds and circles mark the first budgets reaching $P(n)$ ($\beta_C^{\mathrm{match}}$) and the frontier maximum ($\beta_C^{\mathrm{sat}}$), respectively.
}
\label{fig:oracle-frontiers}
\end{figure}

Building on the unified formulation and macro-average frontiers above, we now compare the three strategies through their feasible stopping assignments and budget requirements.
Let $\mathcal S_C$ denote the feasible assignments of stopping positions to all sampled trajectories under strategy $C$.
A zero or sufficiently large shared cap reproduces either routing endpoint, and independent trajectory-level choices can reproduce any shared-cap assignment.
Consequently,
\begin{equation}
\begin{aligned}
\mathcal S_{\mathrm{DR}}&\subseteq\mathcal S_{\mathrm{FB}}\subseteq\mathcal S_{\mathrm{DEE}},
\end{aligned}
\label{eq:oracle-hierarchy1}
\end{equation}
\begin{equation}
\begin{aligned}
F_{\mathrm{DR}}(B)&\leq F_{\mathrm{FB}}(B)\leq F_{\mathrm{DEE}}(B).
\end{aligned}
\label{eq:oracle-hierarchy2}
\end{equation}
An expanded feasible set preserves every previous solution and may admit a more accurate one at the same cost.
The same ordering therefore holds for the macro-averaged frontiers shown in Figure~\ref{fig:oracle-frontiers}.
The DR-to-FB gap reflects the additional potential from allowing question-level intermediate token caps, whereas the FB-to-DEE gap reflects the additional potential from choosing stopping positions separately for sampled trajectories.

Let $A_{\mathrm{full}}=72.33\%$ be the macro-average STP endpoint accuracy $P(n)$, distinct from the original-answer baseline in Table~\ref{tab:appendix-full-thinking}.
We define the smallest relative budget that matches or exceeds this reference as
\begin{equation}
\beta_C^{\mathrm{match}}=\min\bigl\{\beta:\overline F_C(\beta)\geq A_{\mathrm{full}}\bigr\}.
\label{eq:oracle-matching-budget}
\end{equation}
This is where a frontier first reaches the STP full-thinking reference line, marked by a diamond in Figure~\ref{fig:oracle-frontiers}.
The relative budget cap needed to match the STP endpoint reference falls from 71.4\% for dynamic routing to 35.0\% for fixed budgets and 14.2\% for dynamic early exit.

We pair each strategy's maximum accuracy with the first budget that attains it:
\begin{equation}
A_C^{\max}=\overline F_C(1),
\label{eq:oracle-maximum-accuracy}
\end{equation}
\begin{equation}
\beta_C^{\mathrm{sat}}=\min\bigl\{\beta:\overline F_C(\beta)=A_C^{\max}\bigr\}.
\label{eq:oracle-saturation-budget}
\end{equation}
The full-thinking budget accommodates every allowed stopping assignment, so $\overline F_C(1)$ is the measured accuracy ceiling within the strategy's allowed stopping assignments; $\beta_C^{\mathrm{sat}}$ identifies how early that ceiling is reached.
The corresponding macro-averaged measured ceilings are approximately 73\%, 76\%, and 81\% for dynamic routing, fixed budgets, and dynamic early exit, reached at relative budget caps of 94.2\%, 87.2\%, and 58.6\%, respectively.
Thus, within the measured action space and under the macro-averaged frontier construction, trajectory-specific stopping both reaches the STP full-thinking endpoint reference at a lower relative budget cap and attains the highest measured accuracy ceiling at the smallest saturation cap.
These strategy-specific frontiers provide reference curves for evaluating selected public termination methods.

\section{Evaluating Reasoning-Termination Methods with OTBench}

The preceding section constructs Pareto frontiers for oracle methods under the stopping constraints of three termination strategies
To evaluate practical termination methods against these strategy-specific references, we construct \emph{OTBench}, which provides a shared action interface and a dedicated evaluator.
We instantiate OTBench with one publicly available method from each strategy and compare each method's measured token--accuracy curve with the corresponding frontier.

\subsection{A Common Evaluation Protocol for Termination Methods}

For a method $\pi$ operating under strategy $C$ with control parameter $\theta$, let $\mathcal{U}_C$ denote the decision units and let $a_{\pi,\theta}(u)\in\mathcal A_u^C$ be the action selected for unit $u$.
The action is evaluated using the retained token cost $c_u$ and answer accuracy $r_u$ supplied by the STP measurements.
This representation allows methods with different predictors or stopping signals to be evaluated through the same strategy-specific action space.

The average thinking-token cost of method $\pi$ at parameter setting $\theta$ is
\begin{equation}
B_{\pi}(\theta)=
\frac{1}{|\mathcal{U}_C|}
\sum_{u\in\mathcal{U}_C}
c_u\left(a_{\pi,\theta}(u)\right)
\label{eq:method-operating-cost}
\end{equation}
and its corresponding answer accuracy is
\begin{equation}
A_{\pi}(\theta)=
\frac{1}{|\mathcal{U}_C|}
\sum_{u\in\mathcal{U}_C}
r_u\!\left(a_{\pi,\theta}(u)\right).
\label{eq:method-operating-accuracy}
\end{equation}
Here $B_{\pi}(\theta)$ records the mean retained thinking-token cost, while $A_{\pi}(\theta)$ records the mean accuracy of the same selected actions.
The evaluator obtains both quantities from the measured token positions and probe accuracies associated with the selected stopping actions.

For a parameter set $\Theta$, the measured Pareto curve of method $\pi$ is the set of nondominated operating points:
\begin{equation}
\mathcal P_{\pi}
=
\operatorname{ND}
\left\{
\bigl(B_{\pi}(\theta),A_{\pi}(\theta)\bigr):
\theta\in\Theta
\right\},
\label{eq:method-pareto-curve}
\end{equation}
where $\operatorname{ND}$ retains points that are not dominated in both token cost and answer accuracy.
Thus, $\mathcal P_{\pi}$ describes the operating points produced by a concrete method and its scanned control parameter.

OTBench instantiates this evaluation protocol using reasoning trajectories and STP measurements from the above six models across four benchmark groups.
Questions are partitioned into training, validation, and test sets, with all sampled trajectories from the same question assigned to the same partition.
Training measurements supervise learned method components, validation measurements support model selection and parameter calibration, and test measurements are reserved for final evaluation.
On the test set, each method is compared with the frontier induced by its corresponding oracle method using the same questions, aggregation protocol, and token-cost normalization.
The frontiers serve only as evaluation references.
The token-cost metric counts retained thinking tokens and excludes feature-extraction and signal-computation overhead.

\begin{figure}[t]
\centering
\includegraphics[width=\linewidth]{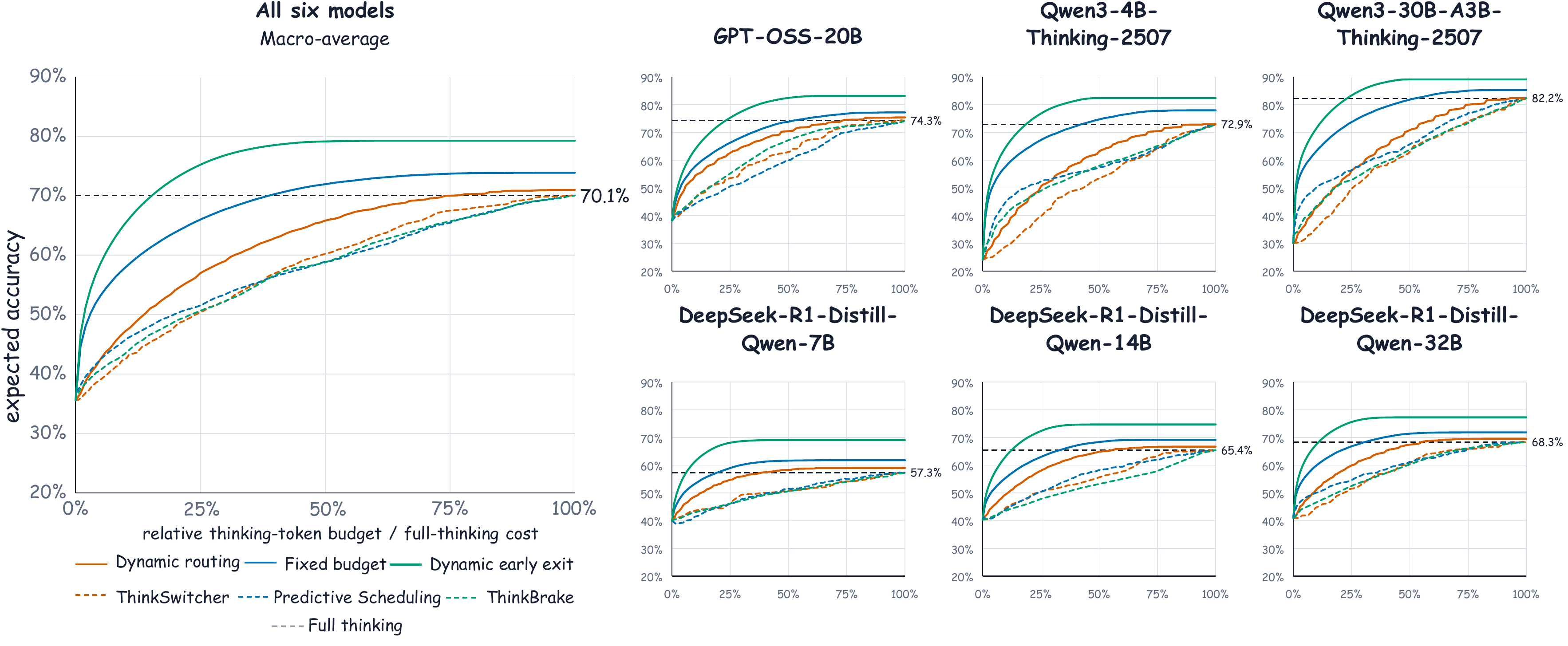}
\caption{
\textbf{Adapted public methods versus strategy-specific oracle frontiers.}
Solid curves show oracle frontiers; same-color dashed curves show the corresponding adapted methods.
The x-axis shows oracle allowed budget caps or public-method realized mean thinking-token costs, each normalized by the corresponding full-thinking token count. 
Horizontal dashed lines mark the macro-average STP full-thinking endpoint accuracy $P(n)$.
}
\label{fig:public-method-frontiers}
\end{figure}

\subsection{Comparing Public Methods with Strategy-Specific Pareto Frontiers}

We adapt one publicly available method from each strategy to the common OTBench interface and evaluate these adaptations on the frozen test measurements.
The adapted ThinkSwitcher~\citep{liang2025thinkswitcher} maps question-level routing to the dynamic-routing endpoints of zero thinking and full thinking.
Predictive Scheduling~\citep{brown2026predictive} outputs a question-level thinking budget that is applied as a shared cap across that question's sampled trajectories.
ThinkBrake~\citep{song2025thinkbrake} evaluates a signal on the current prefix at permitted step boundaries and maps its stop or continue decision to dynamic early exit.
Detailed adaptation and training configurations are provided in Appendix~\ref{app:adaptations}.

We sweep the routing threshold, multiplicative budget scale, and early-exit threshold for ThinkSwitcher, Predictive Scheduling, and ThinkBrake, respectively.
Each parameter setting produces one measured operating point through Equations~\ref{eq:method-operating-cost} and~\ref{eq:method-operating-accuracy}.
The nondominated operating points form the corresponding method curve.
The three curves therefore differ in their learned signals and parameterizations while remaining comparable through the common evaluator.

Figure~\ref{fig:public-method-frontiers} places the three method curves and the frontiers induced by their corresponding oracle methods in a common normalized token--accuracy coordinate system.
Each measured operating point is compared with the corresponding strategy-specific frontier at the same normalized cost, without interpolating additional stopping actions.
This comparison identifies unrealized space relative to the strategy-specific offline references under the evaluated adaptations and frozen STP measurements.

\section{Conclusion}

First, we introduce Step-wise Truncation Probing (STP) as a measurement procedure for characterizing overthinking at the model--task level and show that overthinking is substantial and varies considerably across sampled trajectories.
Then, we formulate dynamic routing, fixed budgets, and dynamic early exit in a unified budget-constrained framework and construct strategy-specific token--accuracy Pareto frontiers as idealized references for their termination potential.
Finally, we develop OTBench, a reusable benchmark built from STP measurements across six models and four benchmark groups.
It contains 70.5B reasoning tokens and provides a common protocol for evaluating practical termination methods against their corresponding strategy-specific frontiers.
Overall, this framework provides a unified way to measure overthinking, quantify termination-strategy potential, and evaluate the performance of practical methods.

\input{references}
\appendix
\clearpage
\input{appendix}

\end{document}

%% file: appendix.tex
\section{Experimental Protocol and Analysis Procedures}
\label{app:protocol}

This appendix specifies the experimental and computational procedures underlying the three main analyses.
We first describe trajectory generation, STP, and answer evaluation, then detail the construction of the oracle frontiers and the adaptations of the public methods.
We conclude with the resampling protocol used for the uncertainty estimates in Appendix~\ref{app:uncertainty}.

\subsection{Models, Tasks, and Generation}

We use six fixed model checkpoints: GPT-OSS-20B, Qwen3-4B-Thinking-2507, Qwen3-30B-A3B-Thinking-2507, DeepSeek-R1-Distill-Qwen-7B, DeepSeek-R1-Distill-Qwen-14B, and DeepSeek-R1-Distill-Qwen-32B.
GPT-OSS-20B has approximately 21B total and 3.6B active parameters, while Qwen3-30B-A3B-Thinking-2507 has 30B total and 3B active parameters; the other checkpoints are dense models.
The frozen configurations identify the corresponding checkpoint repositories and model-specific prompt and channel formats.

The evaluation includes 30 questions from each of AIME2024, AIME2025, and AIME2026, 500 from MATH500, 198 from GPQA-Diamond, and 175 from LiveCodeBench.
The three AIME sets form one 90-question reporting group; the other datasets each form a separate group.
For each of the 963 questions, each model supplies 16 trajectories, giving 92,448 trajectories in total.
Unless otherwise stated, we average trajectories within a question and questions within a benchmark, then give equal weight to the four benchmark groups and six models.
All sampled trajectories enter these summaries.

Generation uses a thinking-token limit of 32,768 and a context limit of 65,536 tokens.
For GPT-OSS-20B, temperature and top-$p$ are both 1.0, top-$k$ filtering is disabled, and the presence penalty is zero.
For Qwen3-4B-Thinking-2507 and Qwen3-30B-A3B-Thinking-2507, temperature is 0.6, top-$p$ is 0.95, top-$k$ is 20, and the presence penalty is 1.5.
For DeepSeek-R1-Distill-Qwen-7B, DeepSeek-R1-Distill-Qwen-14B, and DeepSeek-R1-Distill-Qwen-32B, temperature is 0.6, top-$p$ is 0.95, top-$k$ filtering is disabled, and the presence penalty is zero.
Answer probes use the same model-specific sampling parameters, with a maximum of 4,096 newly generated answer tokens and 16 samples per position.
The inference configuration uses bfloat16, prefix caching, and a GPU-memory-utilization setting of 0.9.

The mathematics instruction requests step-by-step reasoning and a final answer inside \texttt{\textbackslash boxed\{\}}.
For GPQA-Diamond, the instruction requests step-by-step reasoning followed by the choice letter in an answer field; LiveCodeBench uses the step-by-step reasoning instruction.
These instructions occupy the system role for GPT-OSS-20B, Qwen3-4B-Thinking-2507, and Qwen3-30B-A3B-Thinking-2507 and are prepended to the user message for DeepSeek-R1-Distill-Qwen-7B, DeepSeek-R1-Distill-Qwen-14B, and DeepSeek-R1-Distill-Qwen-32B.
Per-sample generation seeds are derived deterministically from the model, benchmark, question, trajectory, and generation phase; STP seeds additionally include the probing position.

Table~\ref{tab:appendix-full-thinking} reports the original full-thinking answer accuracy and mean thinking-token count for each model--benchmark pair.
These accuracies use the original final answers rather than the repeated answer probes used to estimate the STP full-thinking endpoint $P(n)$; corresponding confidence intervals appear in Table~\ref{tab:appendix-full-ci}.

\input{tables/appendix_full_thinking}

\subsection{Probing and Answer Evaluation}
\label{app:stp-protocol}

We construct initial boundaries from tokens containing a double newline and operate on the original thinking-token sequence.
For a trajectory of length $L(n)$, the target number of steps is $\max(1,\lfloor L(n)/b+1/2\rfloor)$, capped by the number of initial blocks, with $b=256$ in the main experiment.
When the initial partition has more steps than this target, we repeatedly merge the adjacent pair with the smallest combined token count, breaking ties in favor of the leftmost pair.
The procedure merges existing blocks without creating new boundaries inside them, so $b$ controls the target resolution rather than imposing a minimum or an exact length on every step.
Appendix~\ref{app:steps} examines the boundary vocabulary and the resulting step-length distributions.

At each position $k=0,\ldots,n$, STP retains the same trajectory prefix and supplies the configured thinking-end marker and answer-channel format.
Qwen3-4B-Thinking-2507, Qwen3-30B-A3B-Thinking-2507, DeepSeek-R1-Distill-Qwen-7B, DeepSeek-R1-Distill-Qwen-14B, and DeepSeek-R1-Distill-Qwen-32B use the \texttt{</think>} delimiter, while GPT-OSS-20B uses its analysis-to-final channel format.
Prompt construction copies the retained token IDs rather than decoding and re-encoding the entire prefix.
At a boundary, the final boundary token is replaced using a precomputed content-token mapping before the thinking-end marker is appended.
The 16 completions at each position are answer generations conditioned on the retained prefix; they are not independently generated reasoning trajectories.
The proportion judged correct defines $P(k)$, including the zero-step and complete-trajectory positions.

The evaluator extracts mathematics answers and checks them against the reference, resolves GPQA choice answers against the reference option, and evaluates extracted code against the supplied LiveCodeBench tests.
An answer for which extraction fails is counted as incorrect, and every probing position retains its denominator of 16 answers.
Correctness is determined by the evaluator rather than by whether a completion reaches its token limit.
The appendix analyses reuse the frozen correctness labels and do not rerun answer generation or relabel the experiments.

The reference position is the earliest maximizer of the measured curve, including ties.
OR uses the fraction of thinking tokens after this trajectory-level measured reference position, while AD uses the difference between its measured accuracy and the STP full-thinking endpoint accuracy $P(n)$.
We distinguish the original full-thinking answer accuracy, represented by one correctness label per trajectory, from the STP full-thinking endpoint $P(n)$, which averages 16 answers generated at the complete-trajectory position.
The full-thinking accuracy interval matrix reports the former; the oracle plots use the latter as their common full-thinking reference.

\subsection{Exact Construction of Strategy-Specific Oracle Frontiers}
\label{app:oracle-computation}

We construct a separate frontier for each model--benchmark cell using the same STP positions, costs, and correctness counts for all three termination strategies.
Dynamic routing assigns one zero-or-full endpoint choice to each question and applies it to all 16 associated trajectories.
Fixed budgets also act at the question level, but enumerate shared absolute token caps from the union of measured token counts across that question's trajectories.
For each cap, every trajectory selects its latest measured position whose retained token count does not exceed the cap.
The cost is the actual retained thinking-token count rather than the nominal cap, and caps selecting the same combination of measured positions produce the same action.
Dynamic early exit instead chooses a measured stopping position separately for each trajectory.

For exact optimization, we represent each action's reward by its integer number of correct probes and its cost by the sum of retained thinking tokens.
Question-level action rewards and costs sum over the question's 16 trajectories, whereas trajectory-level actions contain one trajectory.
We remove dominated actions and combine decision units with a dynamic program that retains the minimum total cost for each attainable total reward.
After combining the units, removing dominated reward--cost states yields the deterministic Pareto frontier.
For any allowed budget, the reported accuracy is the highest attainable reward within that budget divided by the total number of probes, and the selected cost may be strictly smaller than the cap.

Within each cell, the allowed budget cap is normalized by the total token count of the corresponding full-thinking trajectory set.
We evaluate the exact frontiers on a common relative-budget grid and macro-average their accuracies over the four benchmark groups and six models.
This grid is a reporting device and does not determine the admissible actions or replace the exact optimization.
The construction uses neither randomized mixtures of actions nor interpolation between measured stopping positions.
The appendix uncertainty plot displays the original frontier on a 1,001-point grid and its bootstrap intervals on the 101-point grid on which the resampled frontiers were evaluated.
The full-thinking reference is the macro-average of the same cells' $P(n)$ values.
For comparison with public methods, the construction must use only the corresponding test questions and the same test-set reference.

\subsection{Public Methods and Their Adaptations}
\label{app:adaptations}

The main text describes the interface-level mapping of the three public methods.
This section records the adaptation choices needed to implement that mapping and reproduce the corresponding method curves.
The adaptations retain the decision interfaces described in the main text and use question-level partitions, keeping all trajectories of a question together.
ThinkSwitcher~\citep{liang2025thinkswitcher} uses question features to predict the two endpoint accuracies and varies a threshold on their difference.
Its supervision consists of question-level means of the training-set endpoint STP accuracies.
Predictive Scheduling~\citep{brown2026predictive} is adapted to predict a question-level thinking budget and varies a multiplicative scale factor applied to that prediction.
Its specified training target is the mean retained token count at the individual trajectories' $k^\star$ positions; this target differs from the oracle's optimal shared-budget action.
ThinkBrake~\citep{song2025thinkbrake} evaluates the log-probability margin between the most likely non-ending token and the thinking-end token at permitted step boundaries, stopping when the margin is at or below its threshold.

Each parameter setting yields a measured accuracy--cost point, and nondominated points define the method's reported Pareto curve.
Comparisons use measured operating points and the corresponding strategy-specific frontier at a matching allowed cost, without interpolating new operating points.
The cost counts retained thinking tokens, consistent with the main analysis, and does not include predictor training, feature extraction, or signal-computation overhead.

\subsection{Resampling and Uncertainty Estimation}
\label{app:bootstrap}

For the main statistical intervals, we use 10,000 paired two-stage bootstrap replicates with seed 20260721.
Within each reporting benchmark, each replicate samples as many questions as the original group, with replacement, using the same question draw across the six models.
Within each selected model--question occurrence, it then draws 16 of that group's observed trajectories with replacement.
AIME is resampled as one pool of 90 questions.
We recompute the cell summaries and their equal-weight benchmark and model macro-averages within each replicate, and take the 2.5th and 97.5th percentiles as the 95\% interval.
Paired metrics use the same sampled questions and trajectories.

The original correctness label supplies full-thinking accuracy, while the fixed STP curves and token positions supply OR and AD.
The 16-probe curves are held fixed during this bootstrap, so it quantifies question and trajectory resampling uncertainty rather than additional answer-probe uncertainty.
For the frontier intervals, all three deterministic optimization problems are solved again in every replicate on the sampled questions and trajectories, with budgets normalized by the replicate's full-thinking cost.
Percentiles are computed separately at each reported budget; the shaded regions are pointwise intervals rather than simultaneous confidence bands.
The six model identities and four reporting benchmarks are fixed, so these intervals do not describe sampling a broader population of model families or task domains.

Finite-probe and finite-trajectory analyses instead subsample without replacement from the already observed samples, as specified in Appendix~\ref{app:finite-measurement}.
Their conditional variation is distinct from the two-stage bootstrap intervals.

\section{Uncertainty and Measurement Robustness}
\label{app:uncertainty}

This section separates uncertainty in the sampled questions and trajectories from sensitivity to the amount of measurement available for each question.
We first report interval matrices for the principal scalar measurements and pointwise intervals for the oracle frontiers.
We then examine smaller probe and trajectory pools while holding the underlying recorded experiment fixed.

\subsection{Interval Matrices and Oracle Frontiers}

\begin{figure}[!htbp]
\centering
\includegraphics[width=0.85\linewidth]{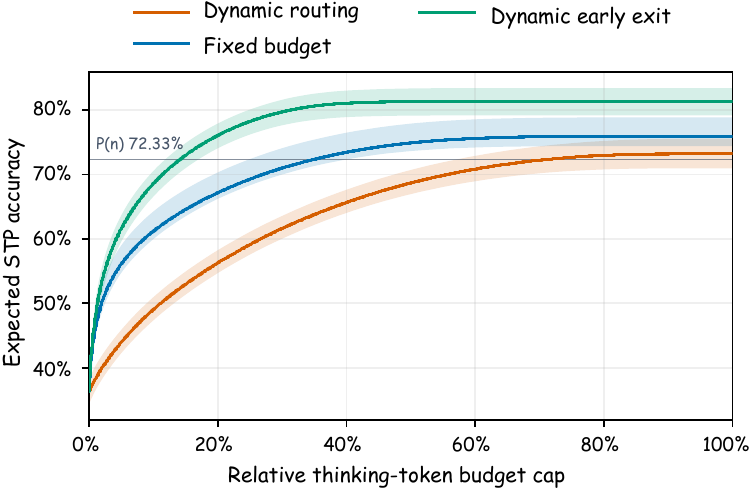}
\caption{
\textbf{Exact deterministic oracle frontiers with pointwise 95\% two-stage bootstrap intervals.}
The analysis uses all six models and four reporting benchmark groups, including the combined AIME set, with equal model and benchmark weights.
The horizontal axis is the allowed budget cap relative to full-thinking cost; shaded regions summarize 10,000 reoptimized bootstrap frontiers at 101 reporting budgets.
The horizontal reference is the macro-average STP full-thinking endpoint accuracy $P(n)$.
Shading does not provide simultaneous coverage of the whole frontier or interpolate new stopping actions.}
\label{fig:appendix-frontier-ci}
\end{figure}

Tables~\ref{tab:appendix-or-ci}--\ref{tab:appendix-full-ci} report intervals only, preserving the model--benchmark matrix layout without repeating point estimates.
Point estimates for OR and AD appear in Table~\ref{tab:main-results}; original full-thinking accuracy and token counts appear in Table~\ref{tab:appendix-full-thinking}.
OR is expressed as a percentage of thinking tokens, AD in percentage points, and original full-thinking answer accuracy in percent.
The Avg. column gives equal weight to the four benchmark groups, and the Avg. row gives equal weight to the six models.
Figure~\ref{fig:appendix-frontier-ci} shows the corresponding pointwise uncertainty for the three strategy-specific oracle frontiers.
Its full-thinking line uses the complete-trajectory STP endpoint, rather than the original-answer accuracy in Table~\ref{tab:appendix-full-thinking}.

\input{tables/appendix_or_intervals}
\input{tables/appendix_ad_intervals}
\input{tables/appendix_full_accuracy_intervals}

\subsection{Sensitivity to Probe and Trajectory Counts}
\label{app:finite-measurement}

We compare 4, 8, 12, and 16 answer probes per stopping position by subsampling the recorded 16 correctness labels without replacement, independently across positions.
For every reduced probe budget, the measured curve and its earliest maximizing position are recomputed before obtaining token-based OR and AD.
We also track whether this maximizing position agrees with the full 16-probe reference, without treating the reference as an infinite-sample optimum.
Separately, we examine 2, 4, 8, and 16 trajectories per model--question by subsampling the recorded trajectory pool without replacement while keeping the questions and probe curves fixed.
All summaries retain flat curves and use the same benchmark and model weights as the main results.
For probe thinning, we integrate the hypergeometric subsampling distribution exactly, including the earliest-maximum tie rule, rather than estimating it with additional generated answers.
At 4, 8, 12, and 16 probes, the expected macro-average OR is 74.69\%, 69.09\%, 66.22\%, and 64.34\%, respectively; expected AD is 11.07, 9.87, 9.34, and 9.03 percentage points.
The corresponding probability of retaining the full-16 earliest maximizer is 51.83\%, 69.94\%, 84.88\%, and 100\%.
Thus reducing probe counts changes the measured reference position and increases the expected reported metrics under this conditional experiment; it does not establish that 16 probes recover an infinite-probe optimum.

For trajectory thinning, we use 2,000 draws with seed 20260721 and report their conditional 2.5th--97.5th percentile ranges.
Uniformly selecting fewer recorded trajectories leaves the conditional expectation of the OR and AD averages unchanged, while widening their conditional sampling variation.
With two trajectories per question, the macro-average ranges are [63.88, 64.82]\% for OR and [8.68, 9.41] percentage points for AD; with eight trajectories, they narrow to [64.16, 64.52]\% and [8.89, 9.17] percentage points.
The full pool of 16 trajectories has no thinning uncertainty because all recorded trajectories are retained.

\begin{figure}[!htbp]
\centering
\includegraphics[width=0.85\linewidth]{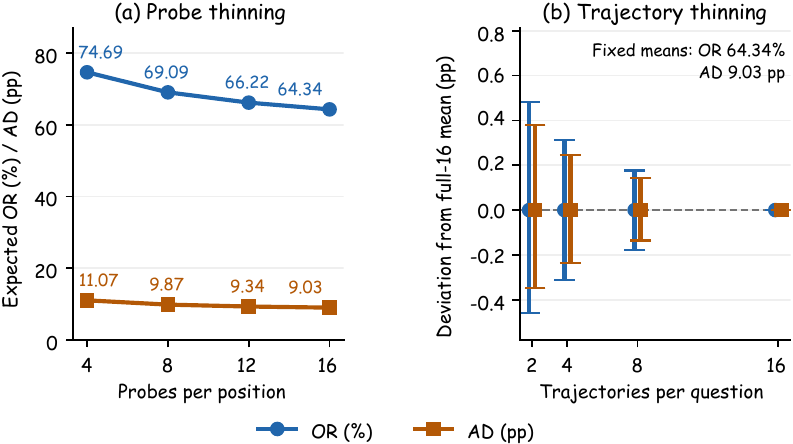}
\caption{
\textbf{Conditional sensitivity to probe and trajectory counts in the frozen experiment.}
Left: OR and AD after reducing the number of probes at each measured position, with the earliest maximum recomputed.
Right: deviations from the full-16 OR and AD means under reduced trajectory pools per model--question, with conditional percentile ranges from 2,000 draws.
Left-panel OR is in percent and AD in percentage points; right-panel deviations are in percentage points.
These conditional analyses keep the observed questions fixed and are distinct from the bootstrap in Appendix~\ref{app:bootstrap}.}
\label{fig:appendix-measurement-sensitivity}
\end{figure}

\subsection{Zero-Step and Direct-Answer Comparisons}
\label{app:zero-step}

The zero-step reference retains no thinking prefix and evaluates answer probes at $k=0$ under the same frozen STP records used elsewhere in the paper.
The resulting artifact contains 5,778 model--question records, corresponding to 963 questions, six models, 16 sampled trajectories per question, and 16 answer probes per trajectory.
For comparison, we also report a separately provided direct-answer condition whose prompt requests an answer without a reasoning trace.
The two conditions are related but not identical: zero-step retains the STP prompt and probe protocol, whereas direct answer uses the separately specified no-thinking prompt.

Table~\ref{tab:appendix-zero-step} reports the descriptive comparison using the same four benchmark groups and equal model and benchmark weights as the main analysis.
The zero-step entries are derived from the frozen STP value $P(0)$, while the direct-answer entries are the returned two-decimal aggregate results.
The overall means are similar descriptively in this model--task set, but this comparison does not establish question-level agreement, statistical equivalence, or the validity of zero-step at every intermediate stopping position.

\input{tables/appendix_zero_step}

\section{Accuracy Curves and Trajectory Diversity}
\label{app:curves}

Aggregate OR and AD describe complementary aspects of the measured trajectories, but do not show how accuracy changes along an individual prefix sequence.
This section provides two kinds of examples: distinct observed curve shapes and differences among trajectories generated for the same model and question.
The examples illustrate the measurements without estimating the population frequency of a curve taxonomy.

\subsection{Representative STP Accuracy Curves}

\begin{figure}[!htbp]
\centering
\includegraphics[width=\linewidth]{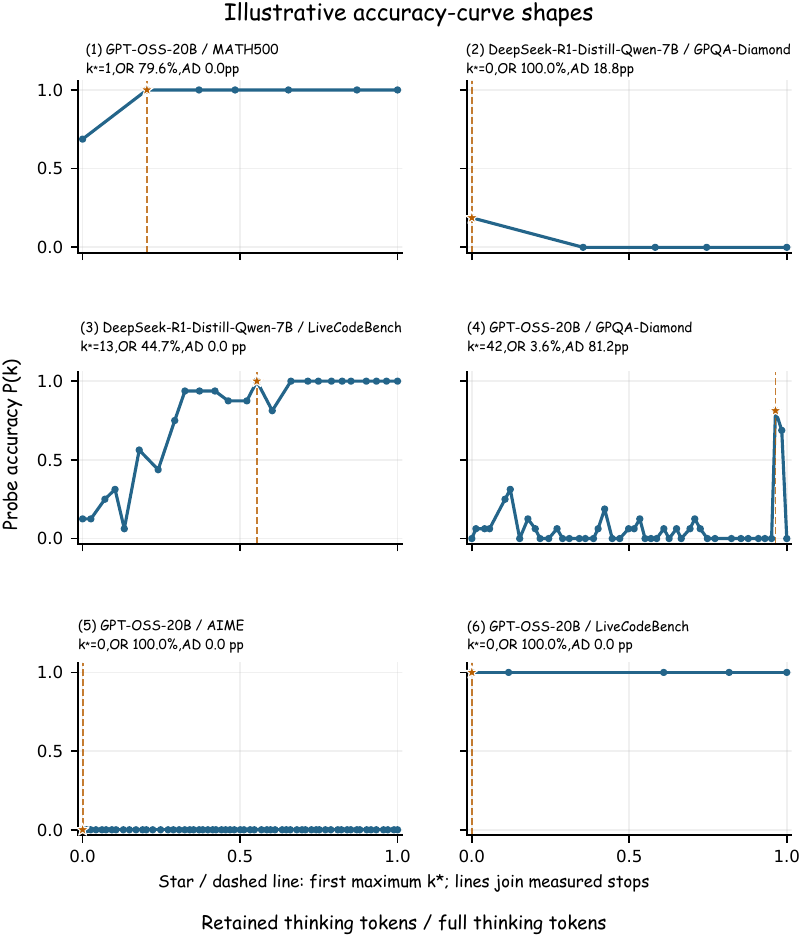}
\caption{
\textbf{Representative measured accuracy curves, shown separately rather than overlaid.}
Each panel identifies its model and benchmark and reports token-based OR and AD; markers locate $k^\star$.
The horizontal coordinate is the retained fraction of the trajectory's full thinking-token count, and the vertical coordinate is $P(k)$.
Line segments connect measured positions for readability and are not additional probe measurements.
The examples illustrate possible shapes and do not estimate their prevalence.}
\label{fig:appendix-curve-shapes}
\end{figure}

Figure~\ref{fig:appendix-curve-shapes} presents recorded STP curves, marking each curve's earliest maximum and reporting its token-based OR and AD.
The curves show how a similar suffix length can coexist with different changes in answer accuracy, and how an early maximum can be followed by a plateau or further fluctuations.
Flat curves follow the same earliest-maximum convention.
All values are empirical proportions from 16 answer probes at each measured position, and a plotted accuracy change need not indicate a change in the model's internal reasoning state.

\subsection{Trajectory Diversity Within a Question}

\begin{figure}[!htbp]
\centering
\includegraphics[width=\linewidth,height=0.73\textheight,keepaspectratio]{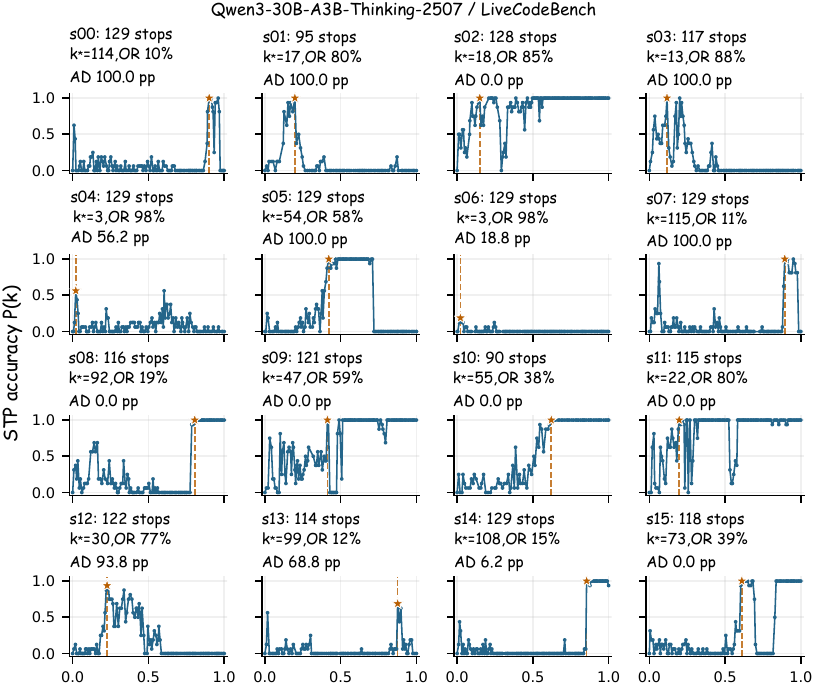}
\caption{
\textbf{Trajectory diversity for one fixed model--question pair.}
Each panel shows one of its 16 sampled trajectories, with the same accuracy and relative thinking-token axes and a marker at its earliest measured maximum.
All trajectories of the selected question are displayed, including any flat curves; no subset is chosen to enlarge the visual contrast.
The displayed model is GPT-OSS-20B and the question is GPQA-Diamond question \texttt{recIj8lR4tuDgrHou}.}
\label{fig:appendix-trajectory-matrix}
\end{figure}

Figure~\ref{fig:appendix-trajectory-matrix} holds the model and question fixed and shows the sampled trajectories in separate panels with shared axis ranges.
This layout permits comparison of accuracy evolution and maximizing positions without obscuring individual curves through overplotting.
The displayed group is GPT-OSS-20B on GPQA-Diamond question \texttt{recIj8lR4tuDgrHou}, selected by a documented deterministic rule for a representative level of within-question variation rather than by maximizing the observed effect.
The example supplements the full-sample pairwise analysis in the main text; it does not establish a frequency or a universal trajectory pattern.

\section{Step Construction and Measurement Resolution}
\label{app:steps}

STP uses operational steps to place probes at controllable resolution.
This section examines the token-level boundary rule and the distribution of resulting step lengths under different measurement resolutions.
The analyses concern the construction and resolution of the measurement used by the main-text local-change analysis.

\subsection{Boundary Identification and Resulting Step Lengths}
\label{app:boundaries}

For each tokenizer, we construct a vocabulary-level set containing every token ID whose individual decoded form contains two consecutive newline characters.
This includes a bare double newline and compound forms in which punctuation, spaces, or other characters accompany it.
After constructing this set, splitting scans the original thinking-token IDs and tests set membership, rather than finding boundaries in a decoded trajectory and mapping them back to tokens.
Thus vocabulary inspection uses decoded token forms, while trajectory segmentation itself operates in token space.
The per-model audit distinguishes boundary occurrences contained inside a token from newline sequences spanning adjacent tokens.
Across all 92,448 trajectories and 576,688,123 stored thinking tokens, the byte-level audit agrees with full-sequence decoding and finds no double-newline pair spanning token boundaries.
The boundary set contains 671 vocabulary IDs for GPT-OSS-20B and 565 for each of the other five models; Table~\ref{tab:appendix-boundary-audit} reports the observed forms separately for each model.
Reconstructing the main $b=256$ partition reproduces every recorded step position, with zero trajectory-level mismatches.
These checks establish compatibility on the recorded experiment rather than a guarantee about arbitrary future model outputs.

\input{tables/appendix_boundaries}

\begin{figure}[!htbp]
\centering
\includegraphics[width=\linewidth]{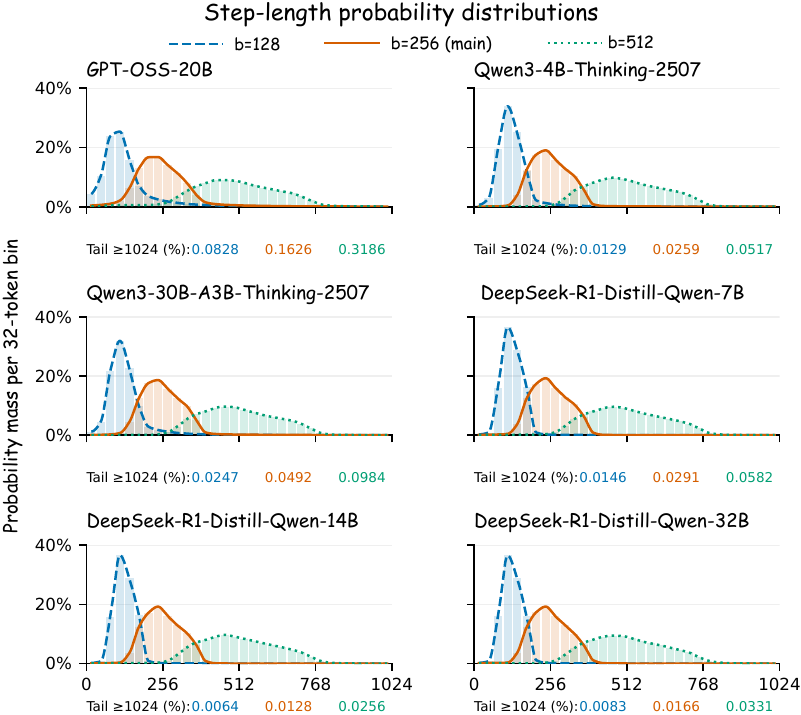}
\caption{
\textbf{Empirical distributions of step-token lengths after repartitioning the same trajectories with $b=128$, $256$, and $512$.}
Panels separate the six models, and $b=256$ denotes the main measurement setting.
The horizontal axis counts tokens per resulting step.
The analysis changes only the partition and does not generate answers at the alternative positions.}
\label{fig:appendix-step-lengths}
\end{figure}

To examine the control supplied by $b$, we repartition the same recorded thinking-token sequences using $b=128$, $256$, and $512$ and leave their generated content unchanged.
Figure~\ref{fig:appendix-step-lengths} shows the empirical distributions of resulting step lengths separately for each model.
Each distribution pools the resulting steps of that model, giving each step equal weight.
The distributions describe realized step lengths, not a lower-bound guarantee or exact per-step length.
Across models, the mean realized step length ranges from 128.18 to 129.92 tokens for $b=128$, from 255.16 to 256.84 for $b=256$, and from 499.80 to 513.01 for $b=512$.
The shift in these distributions makes the role of $b$ as a resolution parameter visible while retaining the variability of natural boundary spacing.
This experiment performs no new answer probing and therefore makes no claim about changes in OR, AD, or oracle frontiers under the alternative partitions.

\clearpage
\section{Limitations and Further Directions}
\label{app:limitations}

\textbf{Offline frontiers are evaluation references rather than deployable stopping policies.}
The frontiers of oracle methods are constructed offline from complete STP curves and optimize measured stopping actions under the structural constraints of each termination strategy.
They therefore quantify the potential available in the measured action space, but do not specify an online policy that can access the same information before the trajectory is complete.

\textbf{STP measurements depend on probing resolution and repeated answer generation.}
The measured accuracy curves depend on the constructed step boundaries, the target average step length, and the number of answer probes at each position.
Probing also requires repeated answer generation at multiple prefixes, while the main token-cost metric counts retained thinking tokens rather than this evaluation overhead.
These choices determine the resolution and measurement uncertainty of OR, AD, and the oracle frontiers.

\textbf{The empirical evaluation covers a fixed set of models, tasks, and method adaptations.}
The main experiments use six models, four benchmark groups, and one adapted public method from each termination strategy.
The reported curves therefore characterize the evaluated settings and do not determine how the same patterns will extend to other model families, task domains, or implementations.

\textbf{Finer-grained and trajectory-aware termination remains an open direction.}
Future work can study policies that use richer information from the generated prefix, support finer-grained stopping decisions, and evaluate a broader range of models and tasks.
Such studies can also incorporate deployment overheads and examine how closely online decisions can approach the offline references.

%% file: tables/appendix_full_thinking.tex
\begin{table}[!htbp]
\centering
\caption{
\textbf{Original full-thinking baselines.}
Acc. is the original full-thinking answer accuracy (\%); Tokens is the mean number of thinking tokens retained by the original full-thinking generations.
These accuracies use the original final answers, not the repeated answer probes used to estimate the STP endpoint $P(n)$.
Avg. uses equal benchmark and model weights.
Benchmarks: AIME combines AIME2024, AIME2025, and AIME2026; MATH: MATH500; GPQA: GPQA-Diamond; LCB: LiveCodeBench.
Models: M1: GPT-OSS-20B; M2: Qwen3-4B-Thinking-2507; M3: Qwen3-30B-A3B-Thinking-2507; M4: DeepSeek-R1-Distill-Qwen-7B; M5: DeepSeek-R1-Distill-Qwen-14B; M6: DeepSeek-R1-Distill-Qwen-32B.
}
\label{tab:appendix-full-thinking}
\begingroup
\fontsize{9}{11}\selectfont
\setlength{\tabcolsep}{1.5pt}
\renewcommand{\arraystretch}{1.12}
\begin{tabular*}{\linewidth}{@{\extracolsep{\fill}}l*{5}{rr}@{}}
\toprule
\multirow{2}{*}{\textbf{Model}}  & \multicolumn{2}{c}{\textbf{AIME}} & \multicolumn{2}{c}{\textbf{MATH}} & \multicolumn{2}{c}{\textbf{GPQA}} & \multicolumn{2}{c}{\textbf{LCB}} & \multicolumn{2}{c}{\textbf{Avg.}} \\
\cmidrule(lr){2-3}\cmidrule(lr){4-5}\cmidrule(lr){6-7}\cmidrule(lr){8-9}\cmidrule(lr){10-11}
 & \textbf{Acc.} & \textbf{Tokens} & \textbf{Acc.} & \textbf{Tokens} & \textbf{Acc.} & \textbf{Tokens} & \textbf{Acc.} & \textbf{Tokens} & \textbf{Acc.} & \textbf{Tokens} \\
\midrule
\textbf{M1} & 75.07 & 8,210 & 93.64 & 1,400 & 65.37 & 3,167 & 72.36 & 7,075 & 76.61 & 4,963 \\
\textbf{M2} & 75.62 & 19,064 & 95.89 & 5,793 & 67.08 & 7,203 & 57.43 & 11,819 & 74.00 & 10,970 \\
\textbf{M3} & 85.28 & 15,649 & 96.04 & 4,716 & 72.66 & 6,756 & 78.04 & 13,004 & 83.00 & 10,031 \\
\textbf{M4} & 48.33 & 12,902 & 93.09 & 3,652 & 51.77 & 6,280 & 42.29 & 11,359 & 58.87 & 8,548 \\
\textbf{M5} & 55.83 & 11,849 & 94.33 & 3,362 & 58.78 & 5,196 & 59.71 & 9,898 & 67.16 & 7,576 \\
\textbf{M6} & 62.85 & 11,049 & 94.84 & 3,142 & 63.51 & 5,404 & 63.00 & 10,799 & 71.05 & 7,598 \\
\midrule
\textbf{Avg.} & \textbf{67.16} & \textbf{13,121} & \textbf{94.64} & \textbf{3,678} & \textbf{63.19} & \textbf{5,668} & \textbf{62.14} & \textbf{10,659} & \textbf{71.78} & \textbf{8,281} \\
\bottomrule
\end{tabular*}
\par
\endgroup
\end{table}

%% file: tables/appendix_or_intervals.tex
\begin{table}[!htbp]
\centering
\caption{
\textbf{Confidence intervals for OR (\%) reported in Table~\ref{tab:main-results}.}
Cells report 95\% percentile intervals from 10,000 paired bootstrap replicates of questions and trajectories, with probe labels fixed.
}
\label{tab:appendix-or-ci}
\begingroup
\footnotesize
\setlength{\tabcolsep}{0pt}
\renewcommand{\arraystretch}{1.12}
\begin{tabular*}{\linewidth}{@{}l@{\extracolsep{\fill}}ccccc@{}}
\toprule
Model & AIME & MATH500 & GPQA-Diamond & LiveCodeBench & Avg. \\
\midrule
\mbox{GPT-OSS-20B} & [30.06, 38.96] & [42.81, 47.91] & [31.65, 39.21] & [45.66, 52.58] & [39.24, 42.87] \\
\mbox{Qwen3-4B-Thinking-2507} & [58.85, 68.10] & [88.03, 90.49] & [68.58, 74.56] & [54.01, 61.17] & [68.81, 72.17] \\
\mbox{Qwen3-30B-A3B-Thinking-2507} & [58.98, 68.32] & [90.89, 93.03] & [70.64, 77.09] & [64.49, 71.77] & [72.69, 76.12] \\
\mbox{DeepSeek-R1-Distill-Qwen-7B} & [58.17, 67.59] & [80.42, 84.20] & [56.01, 61.54] & [59.30, 66.24] & [65.05, 68.38] \\
\mbox{DeepSeek-R1-Distill-Qwen-14B} & [53.88, 64.05] & [77.37, 81.07] & [62.37, 68.06] & [52.15, 59.89] & [63.03, 66.67] \\
\mbox{DeepSeek-R1-Distill-Qwen-32B} & [54.78, 65.00] & [82.77, 86.25] & [60.63, 66.56] & [62.75, 69.65] & [66.78, 70.38] \\
\midrule
Avg. & [53.58, 60.82] & [77.46, 80.08] & [59.43, 63.41] & [57.58, 62.40] & [63.10, 65.60] \\
\bottomrule
\end{tabular*}
\endgroup
\end{table}

%% file: tables/appendix_ad_intervals.tex
\begin{table}[!htbp]
\centering
\caption{
\textbf{Confidence intervals for AD (percentage points) reported in Table~\ref{tab:main-results}.}
Cells report 95\% percentile intervals from 10,000 paired bootstrap replicates of questions and trajectories, with probe labels fixed.
}
\label{tab:appendix-ad-ci}
\begingroup
\footnotesize
\setlength{\tabcolsep}{0pt}
\renewcommand{\arraystretch}{1.12}
\begin{tabular*}{\linewidth}{@{}l@{\extracolsep{\fill}}ccccc@{}}
\toprule
Model & AIME & MATH500 & GPQA-Diamond & LiveCodeBench & Avg. \\
\midrule
\mbox{GPT-OSS-20B} & [2.15, 3.96] & [1.28, 2.08] & [13.44, 18.88] & [8.34, 11.54] & [6.86, 8.51] \\
\mbox{Qwen3-4B-Thinking-2507} & [1.46, 3.76] & [0.68, 1.89] & [11.53, 17.09] & [15.66, 21.09] & [8.05, 10.11] \\
\mbox{Qwen3-30B-A3B-Thinking-2507} & [0.78, 2.42] & [0.98, 2.59] & [8.23, 13.38] & [6.64, 10.80] & [4.81, 6.54] \\
\mbox{DeepSeek-R1-Distill-Qwen-7B} & [7.53, 12.14] & [2.15, 3.54] & [22.57, 28.76] & [12.86, 17.97] & [12.24, 14.59] \\
\mbox{DeepSeek-R1-Distill-Qwen-14B} & [5.76, 9.35] & [1.26, 2.41] & [15.79, 21.81] & [8.09, 11.98] & [8.51, 10.53] \\
\mbox{DeepSeek-R1-Distill-Qwen-32B} & [3.60, 6.42] & [1.15, 2.43] & [13.71, 19.23] & [10.14, 14.83] & [7.92, 9.89] \\
\midrule
Avg. & [4.08, 5.70] & [1.44, 2.28] & [14.84, 19.11] & [10.98, 13.92] & [8.36, 9.72] \\
\bottomrule
\end{tabular*}
\endgroup
\end{table}

%% file: tables/appendix_full_accuracy_intervals.tex
\begin{table}[!htbp]
\centering
\caption{
\textbf{Confidence intervals for original full-thinking answer accuracy (\%) in Table~\ref{tab:appendix-full-thinking}.}
These intervals refer to the original final answers, not STP $P(n)$.
Cells contain only ordinary 95\% percentile confidence intervals from 10,000 paired question-and-trajectory bootstrap replicates; probe labels are fixed.
AIME pools 2024--2026.
The Avg. column weights the four benchmark groups equally; the Avg. row weights models equally.
Aggregate intervals use replicate-level macro-averages, not averages of interval endpoints.
}
\label{tab:appendix-full-ci}
\begingroup
\footnotesize
\setlength{\tabcolsep}{0pt}
\renewcommand{\arraystretch}{1.12}
\begin{tabular*}{\linewidth}{@{}l@{\extracolsep{\fill}}ccccc@{}}
\toprule
Model & AIME & MATH500 & GPQA-Diamond & LiveCodeBench & Avg. \\
\midrule
\mbox{GPT-OSS-20B} & [68.40, 81.39] & [91.89, 95.24] & [60.01, 70.74] & [67.50, 77.00] & [74.17, 79.05] \\
\mbox{Qwen3-4B-Thinking-2507} & [67.78, 82.92] & [94.25, 97.35] & [61.43, 72.66] & [52.25, 62.54] & [71.21, 76.68] \\
\mbox{Qwen3-30B-A3B-Thinking-2507} & [78.89, 91.25] & [94.42, 97.50] & [67.11, 78.03] & [73.32, 82.68] & [80.55, 85.37] \\
\mbox{DeepSeek-R1-Distill-Qwen-7B} & [40.14, 56.53] & [91.21, 94.80] & [46.69, 57.01] & [36.43, 48.43] & [55.97, 61.73] \\
\mbox{DeepSeek-R1-Distill-Qwen-14B} & [47.43, 64.10] & [92.53, 95.91] & [53.19, 64.46] & [53.82, 65.68] & [64.20, 70.10] \\
\mbox{DeepSeek-R1-Distill-Qwen-32B} & [54.24, 71.11] & [93.12, 96.36] & [57.92, 68.94] & [57.18, 68.68] & [68.07, 73.94] \\
\midrule
Avg. & [60.29, 73.72] & [93.17, 95.92] & [58.45, 67.93] & [57.50, 66.70] & [69.37, 74.13] \\
\bottomrule
\end{tabular*}
\endgroup
\end{table}

%% file: tables/appendix_zero_step.tex
\begin{table}[!htbp]
\centering
\caption{\textbf{Zero-step and direct-answer accuracies (\%).} Zero-step retains no thinking prefix and uses the frozen STP answer-probe protocol at $k=0$.
Direct-answer uses a separately reported prompt that requests an answer without a reasoning trace.
Zero-step values are reported to two decimals from the frozen $P(0)$ records; direct-answer values are the returned two-decimal aggregate results.
Models M1--M6 follow Table~\ref{tab:main-results}.
The Avg. column averages the four benchmark groups equally, and the Avg. row averages the six models equally.}
\label{tab:appendix-zero-step}
\begingroup
\footnotesize
\setlength{\tabcolsep}{2pt}
\renewcommand{\arraystretch}{1.12}
\begin{tabular*}{\linewidth}{@{\extracolsep{\fill}}l*{5}{r}@{}}
\toprule
\textbf{Model} & \textbf{AIME} & \textbf{MATH500} & \textbf{GPQA} & \textbf{LCB} & \textbf{Avg.} \\
\midrule
\multicolumn{6}{c}{\textbf{(a) Zero-step}} \\
\midrule
\mbox{M1} & 9.95 & 67.02 & 37.27 & 40.67 & 38.73 \\
\mbox{M2} & 0.60 & 51.74 & 20.08 & 13.88 & 21.57 \\
\mbox{M3} & 1.78 & 59.21 & 32.26 & 22.14 & 28.85 \\
\mbox{M4} & 21.45 & 81.56 & 39.29 & 23.85 & 41.54 \\
\mbox{M5} & 20.42 & 74.98 & 46.63 & 28.36 & 42.60 \\
\mbox{M6} & 24.21 & 80.15 & 46.74 & 25.86 & 44.24 \\
\midrule
\textbf{Avg.} & \textbf{13.07} & \textbf{69.11} & \textbf{37.04} & \textbf{25.79} & \textbf{36.25} \\
\addlinespace[3pt]
\midrule
\multicolumn{6}{c}{\textbf{(b) Direct answer}} \\
\midrule
\mbox{M1} & 8.75 & 67.45 & 41.07 & 42.11 & 39.85 \\
\mbox{M2} & 0.35 & 51.80 & 22.79 & 16.64 & 22.90 \\
\mbox{M3} & 3.13 & 60.46 & 37.15 & 22.79 & 30.88 \\
\mbox{M4} & 19.93 & 79.99 & 39.77 & 20.21 & 39.98 \\
\mbox{M5} & 24.03 & 74.01 & 47.19 & 27.29 & 43.13 \\
\mbox{M6} & 30.42 & 82.58 & 51.13 & 27.17 & 47.83 \\
\midrule
\textbf{Avg.} & \textbf{14.44} & \textbf{69.38} & \textbf{39.85} & \textbf{26.04} & \textbf{37.43} \\
\bottomrule
\end{tabular*}
\endgroup
\end{table}

%% file: tables/appendix_boundaries.tex
\begin{table}[!htbp]
\centering
\caption{
\textbf{Token-boundary audit and decoded forms.}
Vocabulary construction decodes each ID; runtime splitting tests ID membership in $B$ only.
Seen counts observed boundary IDs; Events counts their occurrences.
Cross counts overlapping literal LF--LF pairs crossing token payloads; Unmarked counts those whose right token is outside $B$.
An asterisk denotes incomplete byte/decode verification; unavailable entries are not zeros.
Below, P/W/S/I are vocabulary counts: pure double LF, other whitespace-only, compound double-LF suffix, and compound infix.
The top three observed literal forms are escaped, with occurrence counts; complete forms and reconstruction checks remain in the accompanying CSV/protocol.}
\label{tab:appendix-boundary-audit}
\footnotesize
\begin{tabular*}{\linewidth}{@{}l@{\extracolsep{\fill}}rrrrr@{}}
\hline
Model & $|B|$ & Seen & Events & Cross & Unmarked \\
\hline
\mbox{GPT-OSS-20B} & 671 & 176 & 789,761 & 0 & 0 \\
\mbox{Qwen3-4B-Thinking-2507} & 565 & 182 & 3,393,687 & 0 & 0 \\
\mbox{Qwen3-30B-A3B-Thinking-2507} & 565 & 189 & 2,304,662 & 0 & 0 \\
\mbox{DeepSeek-R1-Distill-Qwen-7B} & 565 & 167 & 3,449,932 & 0 & 0 \\
\mbox{DeepSeek-R1-Distill-Qwen-14B} & 565 & 155 & 3,078,850 & 0 & 0 \\
\mbox{DeepSeek-R1-Distill-Qwen-32B} & 565 & 155 & 3,103,862 & 0 & 0 \\
\hline
\end{tabular*}
\par\medskip
\begin{tabular*}{\linewidth}{@{}l@{\extracolsep{\fill}}p{0.68\linewidth}@{}}
\hline
Model & Vocabulary categories and top three observed forms \\
\hline
\mbox{GPT-OSS-20B} & P/W/S/I: 1/48/552/70.\newline \texttt{".\textbackslash{}n\textbackslash{}n"} (671,962); \texttt{":\textbackslash{}n\textbackslash{}n"} (50,814); \texttt{"\textbackslash{}n\textbackslash{}n"} (20,366). \\
\mbox{Qwen3-4B-Thinking-2507} & P/W/S/I: 1/36/528/0.\newline \texttt{".\textbackslash{}n\textbackslash{}n"} (1,862,578); \texttt{"\textbackslash{}n\textbackslash{}n"} (412,111); \texttt{":\textbackslash{}n\textbackslash{}n"} (377,341). \\
\mbox{Qwen3-30B-A3B-Thinking-2507} & P/W/S/I: 1/36/528/0.\newline \texttt{".\textbackslash{}n\textbackslash{}n"} (1,368,115); \texttt{"\textbackslash{}n\textbackslash{}n"} (245,952); \texttt{":\textbackslash{}n\textbackslash{}n"} (225,673). \\
\mbox{DeepSeek-R1-Distill-Qwen-7B} & P/W/S/I: 1/36/528/0.\newline \texttt{".\textbackslash{}n\textbackslash{}n"} (2,275,688); \texttt{":\textbackslash{}n\textbackslash{}n"} (361,627); \texttt{"\textbackslash{}n\textbackslash{}n"} (309,895). \\
\mbox{DeepSeek-R1-Distill-Qwen-14B} & P/W/S/I: 1/36/528/0.\newline \texttt{".\textbackslash{}n\textbackslash{}n"} (2,028,756); \texttt{":\textbackslash{}n\textbackslash{}n"} (344,263); \texttt{"\textbackslash{}n\textbackslash{}n"} (261,462). \\
\mbox{DeepSeek-R1-Distill-Qwen-32B} & P/W/S/I: 1/36/528/0.\newline \texttt{".\textbackslash{}n\textbackslash{}n"} (2,073,340); \texttt{":\textbackslash{}n\textbackslash{}n"} (346,149); \texttt{"\textbackslash{}n\textbackslash{}n"} (258,028). \\
\hline
\end{tabular*}
\end{table}